\documentclass[conference]{IEEEtran}
\IEEEoverridecommandlockouts
\usepackage{cite}
\usepackage{amsmath,amssymb,amsfonts}
\usepackage{algorithmic}
\usepackage{graphicx}
\usepackage{mathtools}
\usepackage{amsmath,amssymb}
\newlength\mylength
\usepackage{hyperref}
\usepackage{xcolor}
\usepackage{textcomp}
\usepackage{tabularx}
\usepackage{siunitx}
\usepackage{booktabs}
\usepackage{caption}
\DeclareCaptionFont{9pt}{\fontsize{9pt}{10pt}\selectfont}
\usepackage{xcolor}
\usepackage{booktabs}
\usepackage{lipsum}
\usepackage{mathtools}
\usepackage{cuted}
\DeclareMathOperator{\EX}{\mathbb{E}}
\def\BibTeX{{\rm B\kern-.05em{\sc i\kern-.025em b}\kern-.08em
    T\kern-.1667em\lower.7ex\hbox{E}\kern-.125emX}}
\begin{document}

\title{Estimation of Missing Data in Intelligent Transportation System\\
{}
}

\author{\IEEEauthorblockN{ Bahareh Najafi\ \   0000-0002-1030-592X }
\IEEEauthorblockA{\textit{University of Toronto}\\
Toronto, Canada \\
bahareh.najafi@mail.utoronto.ca}
\and
\IEEEauthorblockN{Saeedeh Parsaeefard}
\IEEEauthorblockA{\textit{University of Toronto}\\
Toronto, Canada \\
saeideh.fard@utoronto.ca}
\and
\IEEEauthorblockN{Alberto Leon-Garcia}
\IEEEauthorblockA{\textit{University of Toronto}\\
Toronto, Canada \\
alberto.leongarcia@utoronto.ca}
}\maketitle

\begin{abstract}
Missing data is a challenge in many applications, including intelligent transportation systems (ITS). In this paper, we study traffic speed and travel time estimations in ITS, where portions of collected data are missing due to sensor instability and communication errors at collection points. These practical issues can be remediated by missing data analysis, which are mainly categorized as either statistical or machine learning (ML)-based approaches. Statistical methods require the priori probability distribution of the data which is unknown in our application.  Therefore, we focus on an ML-based approach, Multi-Directional Recurrent Neural Network (M-RNN). M-RNN utilizes both temporal and spatial characteristics of the data. We evaluate the effectiveness of this approach on a TomTom dataset containing spatio-temporal measurements of average vehicle speed and travel time in the Greater Toronto Area (GTA). We evaluate the method under various conditions, where the results demonstrate that M-RNN outperforms existing solutions, e.g., spline interpolation and matrix completion, by up to 58$\%$ decreases in Root Mean Square Error (RMSE).
\end{abstract}


\section{Introduction}
Traffic flow estimation is essential in the ITS in a system implemented at scale in a large city, the accuracy will be highly dependent on how data is collected from diverse sources such as cameras and Bluetooth sensors. This heterogeneity in the source of gathered data along with their different placements and capabilities, e.g., precision, capacity, data sampling rate, image resolution of the cameras, leads variation in spatio-temporal measurements. Moreover, for any type of data collection, the problem of missing data can occur throughout a distributed urban network due to malfunctioning sensors or communication errors among the collection points \cite{shang2018imputation}. Furthermore, full coverage is not provided on some roads due to construction activities, which leads to missing data. Therefore, we need to address this challenge in our application scenario by accurately estimating missing values. Generally, two groups of methods are used for the estimation of missing data: Statistical approaches and ML-based approaches.   Statistical methods such as Bayesian models e.g., Bayesian principal component analysis \cite{bishop1999bayesian}, \cite{nounou2002bayesian} infer the missing points using a learned scheme based on the statistical characteristics of the observed data. They usually assume a prior probability distribution for the observed value and missing data are permuted based on a probability distribution function (PDF). The traffic stream of vehicles is affected by several environmental factors and does not necessarily follow a particular PDF. As for ML-based approaches for estimating missing data,  some research has been conducted on missing data analysis in ITS. Given the ability of RNN-based models, particularly Gated Recurrent Unit (GRU), to capture the temporal long-term dependencies in time series, coupled with dynamic nature of traffic data, GRU is an applicable choice for estimation of missing values in ITS. As such, the authors in \cite{che2018recurrent} incorporate pattern of missing data into RNN for multivariate time series prediction with missing values. Moreover, a bidirectional and unidirectional Long Short Term Memory (LSTM) network has been used in \cite{cui2018deep} to predict network-wide traffic speed for one future timestamp. Previous studies \cite{chung2014empirical} demonstrated the superiority of both LSTM unit and GRU over the traditional RNN unit. Based on their experiments, GRU can outperform LSTM in terms of convergence in CPU time, parameter updates, and generalization. \par

From another perspective, both statistical and ML-based techniques can be roughly divided into prediction-based methods, interpolation-based methods, imputation-based methods, and matrix completion-based methods \cite{yoon2018estimating,shang2018imputation}. Concerning prediction-based methods, for each missing data point, a predicted value is calculated by using the extracted method from historical data \cite{zhong2004estimation}. Interpolation-based methods utilize temporal characteristics of data to estimate the missing values \cite{van2005accurate}, \cite{zhong2004estimation}. On the other hand, imputation-based methods such as \cite{qu2008bpca,shang2018imputation} use spatial characteristics of the data by leveraging traffic data of the adjacent roads or other roads with a similar flow variation within the same time slot to fill in the missing values.
Other recent works, e.g., matrix-completion based methods \cite{tan2013tensor,li2013efficient,ran2016tensor,tan2014low} utilize both the temporal as well as the spatial characteristics in the data to estimate the missing points. 
The method requires synchronized inputs from all sensors \cite{yoon2018estimating}.  This requirement is less likely satisfiable in ITS. Since the average speed of different roads are mostly recorded only at the time of congestion, and not all segments experience the same level of congestion simultaneously.\par
Prediction-based methods do not take advantage of the subsequent sequence of traffic measurements that follow the timestamp where there is a missing value \cite{nounou2002bayesian}.   Interpolation-based and imputation-based methods only utilize temporal or spatial characteristics of the data, respectively, and ignore the benefit of using both dimensions at the same time. In summary, studying missing data analysis in ITS is under-developed. Therefore, the focus of our paper is to select an approach based on the spatio-temporal features of the traffic dataset. As such, we rely on a novel neural network-based architecture called Multi-directional Recurrent Neural Network (M-RNN), inspired by \cite{yoon2018estimating}. The advantages of M-RNN for missing data estimation can be summarized as follows: 1) The architecture exploits the correlation within the time sequences of multivariate features associated with a road segment as well as the correlation among multiple features of the segment through using Bidirectional GRU \cite{chung2014empirical} and fully connected layer, respectively. 2) Bi-GRU captures both forward and backward dependencies in each data stream. 3) The model utilizes deep structure GRU (with several hidden layers) which has a higher level of representation capability of the data stream \cite{graves2013hybrid} \cite{cui2018deep}. 4) The model adopts a mechanism to deal with different sampling rate in case of data heterogeneity. To the best of our knowledge, this is the first time that bi-directional GRU has been used as part of deep learning-based model to estimate the missing multivariate features of traffic data stream in ITS. We examine the results of M-RNN under different conditions and compare the performance to the ideal counterpart solutions. 
The rest of the paper is organized as follows. In subsection~\ref{dataset}, characteristics of our utilized traffic dataset are described and compared to a publicly available medical dataset used in \cite{yoon2018estimating}. We present the problem formulation and our solution approach in subsection~\ref{missrep} and ~\ref{M-RNN2}, respectively. Then, implementation results are illustrated in Sections~\ref{result}, and we conclude in Section~\ref{conclusion}.
\section{System Setup} \label{systemsetup}
In order to facilitate the discussion of our proposed method to our traffic dataset, we first adjust the problem formulation provided in \cite{yoon2018estimating} to ITS. In the rest of the paper, we refer to TomTom traffic dataset \cite{tomtom2015} as TT dataset. TT dataset consists of $N$ road segments. Each road segment is specified by geographic coordinates of its starting and ending points. There exist D streams of multivariate time series-based measurements (average speed and travel time) of length T that consist of a timestamp $t$ and measurements $x$ for each road segment $n$, where $x_t^d(n)$ represents actual measurement of measurement stream $d$ at timestamp $t$ for a particular road segment $n$.  Additionally, $L_d(n)$ is defined as the length of a sequence of data stream $d$ belonging to road segment $n$, where the measurements are conducted.  Each sample is identified as ${x_t^{d_1}(n), x_t^{d_2}(n), ..., x_t^{d_m}(n)}$, where m is the number of features for a road segment. The time interval between two consecutive timestamps are usually not equal ($t_{m+1} - t_m \neq t_{n+1} - t_n$). Timestamps in a sequence are sorted in ascending order $(t_n > t_{n-1})$. Generally, there could be some values in each stream that are not observed and considered as missing. 

\begin{figure*}[!t]
\centering
\includegraphics[scale=0.57]{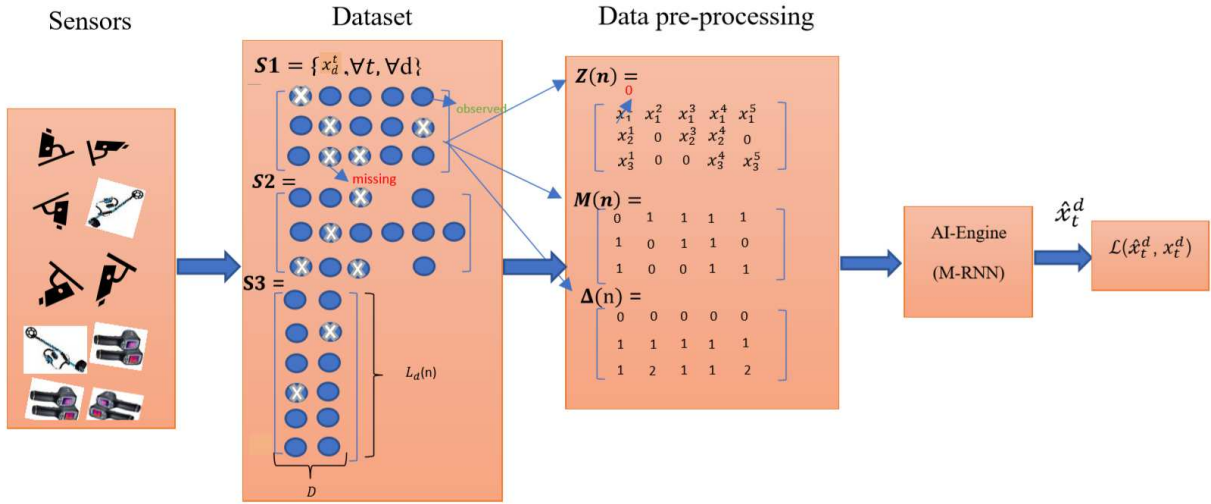}
\caption{System Pipeline for ITS in Our Setup}
\label{fig:fig1111}
\end{figure*}

\subsection{Dataset} \label{dataset}

The TT dataset contains space mean speed \footnote{Space mean speed is defined as the harmonic mean of speeds passing a point during a period of time. It also equals the average speeds over a length of roadway.} \cite{helbing1997fundamentals} and average travel time during peak congestion for several hundreds of approximately one kilometer long, road segments across the Greater Toronto Area. In contrast to most data sources used for traffic parameter estimation, the TT dataset includes the traffic measurements for highways as well as signalized road segments. 
The measurements are collected on a minute basis during congestion, where the average space mean speed is less than $70\%$ of vehicular speed during free-flow conditions. Therefore, the collected per-minute measurements do not contain a value for every road segment.
 We select a three-hour interval from 4:30pm to 7:30pm on 8 September 2017 which is expected to have a high proportion of collected records. During this interval, we select road segments with a significant number of mutual timestamps where measurements for a sufficient number of the segments are conducted. There should be a trade-off among the number of road segments $N$ and $L_d(n)$. Groups of neighbouring road segments in a traffic network are usually influenced by each other in a short period and thus, experience higher number of mutual congestion intervals. Road segments that are far apart from centres of congestion are less likely to share the same congestion period with the underlying segments and consequently have a lower potential to be selected as samples for our experiment. Therefore, the higher the $L_d(n)$ of mutually congested interval, the lower the number of road segments $N$. \par
There are two strongly correlated features in the TT dataset: average speed (${d_1}$) and travel time (${d_2}$). M-RNN takes advantage of the correlation between the time values of a particular data stream $d$ associated with a road segment $n$, as well as the correlation across the streams (${d_1}$) and (${d_2}$) of the underlying road $n$, where $L_{d_{1}}(n)=L_{d_{2}}(n)=L_d(n)$. We demonstrate the performance of our results with varying numbers of samples $N$ and $L_d(n)$. We assume that data is Missing at Random (MAR), which is a reasonably valid assumption in our scenario. In the case of MAR, the missingness mechanism does not depend on unobserved data, and thus the underlying mechanism can be ignored when estimating these values \cite{rubin1976inference}.\par

\subsection{Missing Data Presentation and Estimation} \label{missrep}

Our system pipeline is illustrated in Fig.~\ref{fig:fig1111}. Data is collected from several sensors, e.g., loop detectors \cite{gerlough1976traffic}, cameras on a minute basis. Next, we sort the data into a series of two-dimensional arrays. Each array presents measurements that correspond to a road segment $n$. Each row consists of $x_t^{d_1}, x_t^{d_2}, .., $ and each column shows $x_{t^\prime}^d(n)$, where $t^\prime = 1, 2, .., L_d(n)$. \par In general, the collected data may come from heterogeneous sensors, and the corresponding measurements can be recorded on a different aggregation level. Therefore, we demonstrate three forms of two-dimensional arrays to highlight that the method is capable of handling all the cases. $\textbf{S1}$ can be a representative of homogeneous data that consists of a large number of $D$ and low-frequency measurements (a small number of $L_D(n)$). The number of $D$ and $L_D(n)$ depicted  in Fig.~\ref{fig:fig1111} does not indicate the actual length of features and time series, respectively, and is just used for comparison among the three instances. $\textbf{S2}$ indicates heterogeneous data where the records for different stream of a road segment $n$ may come from multiple sources and subsequently with different sampling frequencies. The third instance, $\textbf{S3}$, demonstrates homogeneous data with small $D$ and large $L_D(n)$, compared to $\textbf{S1}$ and $\textbf{S2}$. Our study includes $\textbf{S3}$. Moreover, the multiplication sign inside the circle in box 2 of Fig.~\ref{fig:fig1111} indicates missing data points for all three instances of input arrays. 

\par The underlying extracted arrays are the input to the pre-processing block.  In this step, we need to produce three arrays in order to use as an input to the AI engine: 
$Z(n),$ $M(n)$ and $\Delta(n)$. $Z(n)$ consists of $x_t^d(n)$ for $d = 1, 2, .., D$ and $t= 1, 2, .., L_d(n)$  for each road segment $n$. We set $x_t^d$ equal to zero, where the data point is missing. $M(n)$ is an array containing 0s and 1s that indicate the coordinates of missing and observed records, respectively, corresponding to array $Z(n).$  As for  $\Delta(n)$, each element of the array, $\delta_t^d(n)$, is defined to be the actual time passed from timestamp $t$ since the last measurement is conducted for each road segment $n$.   More detailed information on how to calculate $\delta_t^d(n)$ can be found in \cite{yoon2018estimating}. The $\Delta(n)$ is defined to handle different sampling rate associated with data heterogeneity. The triplet of generated arrays will then proceed to the AI engine where the M-RNN will be applied, the missing records will be estimated, and the respective RMSE will be calculated. The entire dataset $\mathcal{D}$ consists of the sequence of timestamps and the streams of measurements for all road segments that are sampled from an unknown distribution $\mathcal{F}$, and is mathematically represented as follows in equation \ref{eq:222}.
\begin{equation} \label{eq:222}
\begin{aligned}
\mathcal{D}=\sum_{n = 1}^{N} \{\mathcal{X}(n), \mathcal{S}(n)\},
\end{aligned}
\end{equation}
where $\mathcal{X}(n)=\{x_t^d(n), \forall d, \forall t, \forall n \}$ and $\mathcal{S}(n) = \{t1, t2, ..., L_d(n), \forall n\}$. \par
Our goal is to find the best estimate ($\hat{x}_t^d$), with minimum RMSE for a particular missing measurement at time $t$, where the estimate form and loss function $\mathcal{L} (x_t^d , \hat{x}_t^d)$ to be defined as follows:
\begin{equation}
\begin{split}
& \hat{x}_t^d = f_t^d(\mathcal{X}, \mathcal{S})  \\
& \mathcal{L} (x_t^d , \hat{x}_t^d) = (\hat{x}_t^d - x_t^d)^2
\end{split}
\end{equation}
Then, the optimization problem is to find the function $f_d^t$ that solves the following equation \cite{yoon2018estimating}.
\begin{eqnarray}
&& \min_{\text{f}} \EX_{\mathcal{F}} \left[ \sum_{t = 1}^{T} \sum_{d = 1}^{D} (1 - m_t^d) \mathcal{L} (\mathcal{X}_t^d , \widehat{\mathcal{X}}_t^d)  \right] \nonumber\\
&&= \min_{\text{f}} \EX_{\mathcal{F}} \left[ \sum_{t = 1}^{T} \sum_{d = 1}^{D} (1 - m_t^d) (x_t^d - f_t^d(\mathcal{X} , \mathcal{S}))^2 \right],
\end{eqnarray}

where $\text{f}, \mathcal{F}, \hat{x}$ are desired estimator function, unknown distribution which the measurements are sampled from, and the estimated value for the missing point respectively. Besides, $m$ is an element of the array $\mathbf{M}$ which is defined in the pre-processing box of Fig.~\ref{fig:fig1111}.

\subsection{Multi-directional Recurrent Neural Network (M-RNN)} \label{M-RNN2}
Our objective is to minimize the RMSE of each estimated value $\hat{x}_t^d$ with respect to the missing value $x_t^d$ for each road segment $n$ that is sampled from TT dataset.
In contrast to interpolation-based and imputation-based methods, M-RNN utilizes the correlation from within the stream and the correlation across the streams of average speed and travel time in TT dataset. \par
\begin{figure*}[!t]
\centering
\includegraphics[scale=0.6]{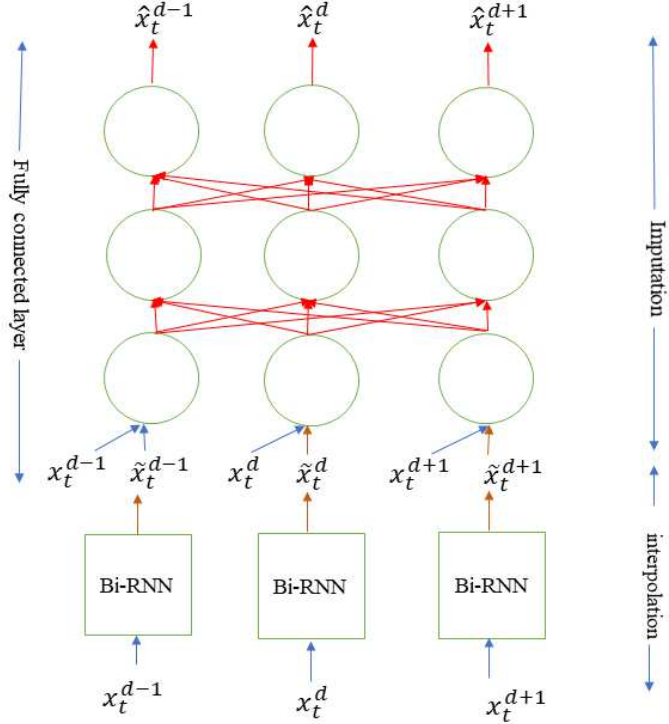}
\caption{M-RNN architecture in time domain \protect\cite{yoon2018estimating}}
\label{fig:fig2}
\end{figure*}

 Fig.~\ref{fig:fig2} illustrates the M-RNN architecture in the time domain. It comprises of one interpolation block and one imputation block which the former comes before the latter in order to improve the accuracy of estimation \cite{yoon2018estimating}. The input to interpolation block consists of feature vector $x$, which are individual columns of array $Z$ explained in section \ref{missrep}. The output of the block (${x}_t^d, \forall d $) is built using a Bi-GRU, where its input to forward, and backward hidden states comes from t – 1 and t + 1 respectively. The imputation block is constructed by one fully connected layer; both blocks are trained simultaneously and the output of imputation block, $\hat{x}_t$ is the final estimate of the missing measurement at time $t$.
Estimation of missing values using M-RNN can be implemented in two ways: 1) single imputation, 2) multiple imputations. Imputing data points using multiple imputations is applied to reduce uncertainty by generating multiple imputed values. Due to the negligible uncertainty level in our data, and in order to reduce complexity, we select to run our experiments under single imputation. Once $\hat{x}_t^d$ is estimated, the RMSE will be calculated through $\mathcal{L} (x_t^d , \hat{x}_t^d)$. The total loss for the entire dataset $\mathcal{D}$ is calculated based on equation \ref{eq:232} \cite{yoon2018estimating}, where $L_n$ and  $m_{t}^{d}(n)$ are the length of time sequence for road segment $n$, and an element of the mask array $\mathbf{M}$, corresponding to timestamp $t$ and road $n$, respectively.

\begin{eqnarray}\label{eq:232}
&&\mathcal{L}(\{ \widehat{\boldsymbol{X}}_{t}^{d}, \boldsymbol{X}_{t}^{d} \})=\nonumber \\
&&\sum_{n=1}^{N} \left[ \frac{ \sum_{t=1}^{L_n}\sum_{d=1}^{D} m_{t}^{d}(n) \times (\hat{x}_{t}^{d}(n) - x_{t}^{d}(n))^2 }
{\sum_{t=1}^{L_n}\sum_{d=1}^{D} m_{t}^{d}(n) }\right] 
\end{eqnarray}

\section{Performance Evaluation} \label{result}
We compare the performance result of M-RNN with single imputation, with two other methods: Spline interpolation and matrix completion of \cite{mazumder2010spectral}. Table~\ref{tab:ParamValues} illustrates the hyperparameters that are optimized in our experiments. We use RMSE as a measure of estimation accuracy of the missing values on TT dataset. In the pre-processing step, we select to remove missing points associated to each feature independently according to probability $\tau$, also named as missing threshold. Next, we apply it to all road segments under experiment.\par
Before moving forward, it is worthwhile to compare the datasets in  \cite{yoon2018estimating}, where the M-RNN is initially applied, and TT dataset. In \cite{yoon2018estimating}, the authors applied their model on several publicly available medical datasets, including MIMIC-III \cite{johnson2016mimic} and Deterioration  \cite{alaa2017personalized}. In terms of data collection method, representation model, and temporal condition, there are differences between the TT dataset and the datasets in \cite{yoon2018estimating}. The TT is a complete dataset consisting of historical aggregated (per minute) time-series-based measurements, whereas the medical datasets have a lower level of aggregation, where features are measured less frequently, every few weeks. Moreover, there is a significant level of sparsity in these medical datasets; many measurements were not conducted, which consequently leads to high missing rates.
Medical datasets, on the other hand, have an enormous number of samples. 20000-30000 patients are involved in their experiments. Each patient has between 20-40 recorded features that are far larger than the two features in our case. However, we note some additional features might exist in other traffic datasets, e.g., time mean speed, traffic flow, density \cite{may1990traffic}, \cite{helbing1997fundamentals}, occupancy percentile \cite{may1990traffic}, volume per lane per mile, and may improve the estimation accuracy of missing points.
Another difference between the medical datasets and TT dataset is the level of uncertainty. There is a high amount of uncertainty in medical measurements \cite{yoon2018estimating} due to individual, systematic and clinical factors. In the TT dataset, the uncertainty level is negligible.  There exists a feature in our dataset that indicates the quality of each record, given as a percentage. All utilized instants in our study have values higher than $95\%$ of the underlying feature.  

\begin{table}
    \begin{center} 
    \caption{Parameters and optimized hyperparameters used in our experiments}
    \label{tab:ParamValues}
    \begin{tabular}{l p{0.85cm}}
    \toprule
    Parameter         & Value   \\
    \midrule  
     Number of road segments (samples)                                         &   382 \\
    \midrule
     Number of data streams                                                    &   2 \\
    \midrule
     Length of time sequence                                                    &   85 \\
    \midrule
     Optimization method                                                        &   Adam \cite{kingma2014adam} \\
    \midrule
     Depth of the network                                                       &   4 \\
    \midrule
     Number of layers per block                                                &   2 \\
    \midrule
     Number of hidden nodes per each layer in interpolation block              &   2 \\
     \midrule
     Number of hidden nodes per each layer in imputation block                 &   2 \\
    \bottomrule
    \end{tabular}
    \end{center} 
\end{table}

\begin{table}
    \begin{center} 
    \caption{Performance evaluation of M-RNN vs. counterpart methods}
    \label{tab:comparisons}
    \begin{tabular}{cccc}
    \hline
    Category     &   Algorithm        &     $RMSE_x$          & $\hat{\eta}_{M-RNN/X}$ $(\%)$ \\
    \hline  
     M-RNN      &   M-RNN \cite{yoon2018estimating}           & 0.01278       & -  \\
    \hline
    Interpolation & Spline              & 0.01938       & 51.64 \\
    \hline
    Matrix completion & \cite{mazumder2010spectral}  & 0.02021 & 58.13  \\
    \hline
    \end{tabular}
    \end{center} 
\end{table}

\par
\begin{figure}[t!]
\centering
\includegraphics[width=0.45\textwidth]{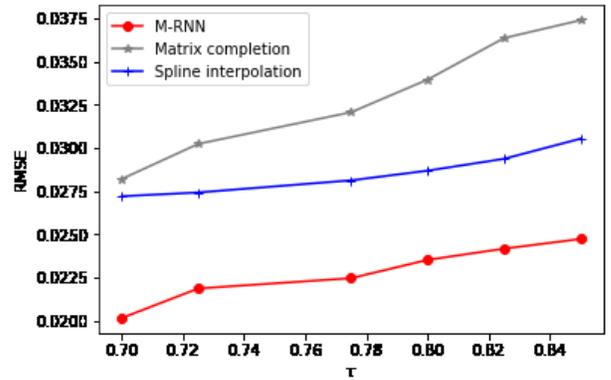}
\caption{Estimation accuracy versus missing threshold ($\tau$)}
\label{fig:figa}
\end{figure}

\begin{figure}[t!]
\centering
\includegraphics[width=0.45\textwidth]{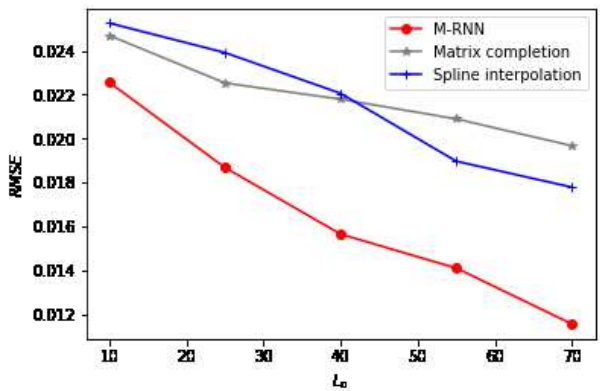}
\caption{Estimation accuracy versus number of measurements per segment ($L_n$)}
\label{fig:figc}
\end{figure}

\begin{figure}[t!]
\centering
\includegraphics[width=0.45\textwidth]{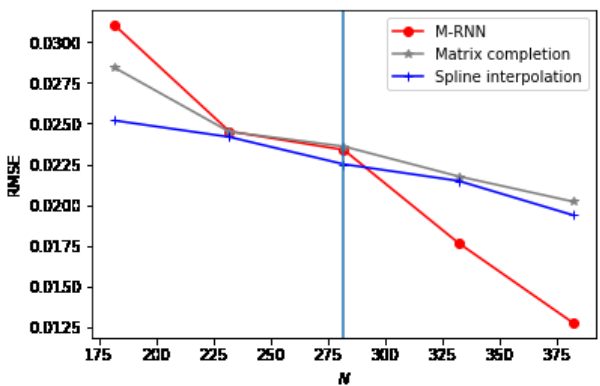}
\caption{Estimation accuracy versus number of road segments ($N$)}
\label{fig:figd}
\end{figure}

 Table~\ref{tab:comparisons} shows the average RMSE for all three methods, where we use 5 cross-fold validation. We have defined $\hat{\eta}_{M-RNN/X}$ in equation~\ref{eq:3} as a means to calculate the improvement made by M-RNN approach in terms of RMSE reduction compared to the existing method. 
\begin{eqnarray}\label{eq:3}
&& \hat{\eta}_{ M-RNN/X} = \frac{\left|RMSE_{M-RNN} - RMSE_X\right|}{RMSE_{M-RNN}} \times 100,
\end{eqnarray}
where $\left| \ast \right|$ indicates absolute value of $\ast$.
From table~\ref{tab:comparisons}, M-RNN significantly outperforms the other two methods.The second-best performer is the Spline interpolation, since TomTom has frequent measurements coupled with a strong correlation among features. We do not compare M-RNN to imputation-based methods, since these methods perform well in the presence of high dimensional data, and there are only two features in TT dataset.
We have conducted further experiments to examine the proposed method under two probability distribution functions of missing data, increasing variance of missing points under Gaussian distribution, increasing the missing threshold ($\tau$), decreasing the number of $L_D(n)$ per road segment $n$ and decreasing the number of $N$. The results are illustrated in figures~\ref{fig:figa}-~\ref{fig:figd}. \par 
Fig.~\ref{fig:figa} illustrates the performance of M-RNN for various values of $\tau$. $\tau$ indicates the probability for a data point to be marked as missing. As can be seen in the plot, M-RNN is remarkably robust to the changes in $\tau$. As $\tau$ increases, the rate of improvements of RMSE in M-RNN ($\hat{\eta}_{\scriptscriptstyle M-RNN/X}$) increases compared to Spline interpolation. \par
 Fig.~\ref{fig:figc} reveals the evaluation result of the model versus number of $L_d(n) = L$. It is highlighted that the performance of M-RNN downgrades as the number of $L_d(n)$ decreases, since the interpolation block of M-RNN in Fig.~\ref{fig:fig2} exploits the correlation among successive measurements associated with a single road. Therefore, the underlying block performs more effectively in the presence of a high sampling rate.\par Fig.~\ref{fig:figd} illustrates the functionality of our trained model under different numbers of road segments ($N$). The RMSE is an ascending function of number of segments $N$, as the method requires sufficient number of samples to learn the parameters. From Fig.~\ref{fig:figd}, for $\mathbf{n} < 275$ (vertical line), both Spline interpolation and matrix completion outperform M-RNN. It is noted that the underlying number depends on the unique characteristics of the dataset, including the number $D$ in conjunction with the intra-stream and inter-stream correlation plus the number of samples $N$ and more importantly, the way we optimize the model.

\section{Conclusion} \label{conclusion}
Traffic flow estimation is of high importance in intelligent transportation system. However, the challenge of missing data is inevitable for all data collection methods due to several reasons. Therefore, an accurate estimation of missing values is essential so that we can take full advantage of our data. In this paper, we exploit a deep learning based approach, M-RNN, for estimating missing data points on TomTom traffic dataset that consists of spatio-temporal features for several hundred segments in GTA. The utilized method takes advantage of inter-stream and intra-stream correlations in the data. Several experiments of the method under various conditions demonstrate significant performance improvement in terms of RMSE reduction, compared to Spline interpolation and matrix completion. We will utilize M-RNN on other ITS datasets, and also extend our work to address cases, where data is missing not at random (MNAR) in future works.

\bibliographystyle{IEEEtran}
\bibliography{myreferences}

\begin{thebibliography}{10}
\providecommand{\url}[1]{#1}
\csname url@samestyle\endcsname
\providecommand{\newblock}{\relax}
\providecommand{\bibinfo}[2]{#2}
\providecommand{\BIBentrySTDinterwordspacing}{\spaceskip=0pt\relax}
\providecommand{\BIBentryALTinterwordstretchfactor}{4}
\providecommand{\BIBentryALTinterwordspacing}{\spaceskip=\fontdimen2\font plus
\BIBentryALTinterwordstretchfactor\fontdimen3\font minus
  \fontdimen4\font\relax}
\providecommand{\BIBforeignlanguage}[2]{{%
\expandafter\ifx\csname l@#1\endcsname\relax
\typeout{** WARNING: IEEEtran.bst: No hyphenation pattern has been}%
\typeout{** loaded for the language `#1'. Using the pattern for}%
\typeout{** the default language instead.}%
\else
\language=\csname l@#1\endcsname
\fi
#2}}
\providecommand{\BIBdecl}{\relax}
\BIBdecl

\bibitem{shang2018imputation}
Q.~Shang, Z.~Yang, S.~Gao, and D.~Tan, ``An imputation method for missing
  traffic data based on fcm optimized by pso-svr,'' \emph{Journal of Advanced
  Transportation}, vol. 2018, 2018.

\bibitem{bishop1999bayesian}
C.~M. Bishop, ``Bayesian pca,'' in \emph{Advances in neural information
  processing systems}, 1999, pp. 382--388.

\bibitem{nounou2002bayesian}
M.~N. Nounou, B.~R. Bakshi, P.~K. Goel, and X.~Shen, ``Bayesian principal
  component analysis,'' \emph{Journal of Chemometrics: A Journal of the
  Chemometrics Society}, vol.~16, no.~11, pp. 576--595, November 2002.

\bibitem{che2018recurrent}
Z.~Che, S.~Purushotham, K.~Cho, D.~Sontag, and Y.~Liu, ``Recurrent neural
  networks for multivariate time series with missing values,'' \emph{Scientific
  reports}, vol.~8, no.~1, pp. 1--12, 2018.

\bibitem{cui2018deep}
Z.~Cui, R.~Ke, Z.~Pu, and Y.~Wang, ``Deep bidirectional and unidirectional lstm
  recurrent neural network for network-wide traffic speed prediction,''
  \emph{arXiv preprint arXiv:1801.02143}, 2018.

\bibitem{chung2014empirical}
J.~Chung, C.~Gulcehre, K.~Cho, and Y.~Bengio, ``Empirical evaluation of gated
  recurrent neural networks on sequence modeling,'' \emph{arXiv preprint
  arXiv:1412.3555}, 2014.

\bibitem{yoon2018estimating}
J.~Yoon, W.~R. Zame, and M.~van~der Schaar, ``Estimating missing data in
  temporal data streams using multi-directional recurrent neural networks,''
  \emph{IEEE Transactions on Biomedical Engineering}, vol.~66, no.~5, pp.
  1477--1490, March 2018.

\bibitem{zhong2004estimation}
M.~Zhong, P.~Lingras, and S.~Sharma, ``Estimation of missing traffic counts
  using factor, genetic, neural, and regression techniques,''
  \emph{Transportation Research Part C: Emerging Technologies}, vol.~12, no.~2,
  pp. 139--166, April 2004.

\bibitem{van2005accurate}
J.~Van~Lint, S.~Hoogendoorn, and H.~J. van Zuylen, ``Accurate freeway travel
  time prediction with state-space neural networks under missing data,''
  \emph{Transportation Research Part C: Emerging Technologies}, vol.~13, no.
  5-6, pp. 347--369, October 2005.

\bibitem{qu2008bpca}
L.~Qu, Y.~Zhang, J.~Hu, L.~Jia, and L.~Li, ``A bpca based missing value
  imputing method for traffic flow volume data,'' in \emph{2008 IEEE
  Intelligent Vehicles Symposium}.\hskip 1em plus 0.5em minus 0.4em\relax IEEE,
  June 2008, pp. 985--990.

\bibitem{tan2013tensor}
H.~Tan, G.~Feng, J.~Feng, W.~Wang, Y.-J. Zhang, and F.~Li, ``A tensor-based
  method for missing traffic data completion,'' \emph{Transportation Research
  Part C: Emerging Technologies}, vol.~28, pp. 15--27, March 2013.

\bibitem{li2013efficient}
L.~Li, Y.~Li, and Z.~Li, ``Efficient missing data imputing for traffic flow by
  considering temporal and spatial dependence,'' \emph{Transportation research
  part C: emerging technologies}, vol.~34, pp. 108--120, September 2013.

\bibitem{ran2016tensor}
B.~Ran, H.~Tan, Y.~Wu, and P.~J. Jin, ``Tensor based missing traffic data
  completion with spatial--temporal correlation,'' \emph{Physica A: Statistical
  Mechanics and its Applications}, vol. 446, pp. 54--63, March 2016.

\bibitem{tan2014low}
H.~Tan, J.~Feng, Z.~Chen, F.~Yang, and W.~Wang, ``Low multilinear rank
  approximation of tensors and application in missing traffic data,''
  \emph{Advances in Mechanical Engineering}, vol.~6, p. 157597, February 2014.

\bibitem{graves2013hybrid}
A.~Graves, N.~Jaitly, and A.-r. Mohamed, ``Hybrid speech recognition with deep
  bidirectional lstm,'' in \emph{2013 IEEE workshop on automatic speech
  recognition and understanding}.\hskip 1em plus 0.5em minus 0.4em\relax IEEE,
  2013, pp. 273--278.

\bibitem{tomtom2015}
H.~T.~H. TomTom, ``Traffictm and iq routestm data provides the very best
  routing,'' \emph{White Paper}, 2010.

\bibitem{helbing1997fundamentals}
D.~Helbing, ``Fundamentals of traffic flow,'' \emph{Physical Review E},
  vol.~55, no.~3, p. 3735, March 1997.

\bibitem{rubin1976inference}
D.~B. Rubin, ``Inference and missing data,'' \emph{Biometrika}, vol.~63, no.~3,
  pp. 581--592, December 1976.

\bibitem{gerlough1976traffic}
D.~L. Gerlough and M.~J. Huber, ``Traffic flow theory,'' Tech. Rep., 1976.

\bibitem{mazumder2010spectral}
R.~Mazumder, T.~Hastie, and R.~Tibshirani, ``Spectral regularization algorithms
  for learning large incomplete matrices,'' \emph{Journal of machine learning
  research}, vol.~11, no. Aug, pp. 2287--2322, August 2010.

\bibitem{johnson2016mimic}
A.~E. Johnson, T.~J. Pollard, L.~Shen, H.~L. Li-wei, M.~Feng, M.~Ghassemi,
  B.~Moody, P.~Szolovits, L.~A. Celi, and R.~G. Mark, ``Mimic-iii, a freely
  accessible critical care database,'' \emph{Scientific data}, vol.~3, p.
  160035, May 2016.

\bibitem{alaa2017personalized}
A.~M. Alaa, J.~Yoon, S.~Hu, and M.~Van~der Schaar, ``Personalized risk scoring
  for critical care prognosis using mixtures of gaussian processes,''
  \emph{IEEE Transactions on Biomedical Engineering}, vol.~65, no.~1, pp.
  207--218, April 2017.

\bibitem{may1990traffic}
A.~D. May, \emph{Traffic flow fundamentals}, 1990.

\bibitem{kingma2014adam}
D.~P. Kingma and J.~Ba, ``Adam: A method for stochastic optimization,''
  \emph{arXiv preprint arXiv:1412.6980}, 2014.

\end{thebibliography}
\vspace{12pt}
\color{red}

\end{document}